%% file: arxiv_version.tex
\documentclass[10pt,twocolumn,letterpaper]{article}

\usepackage{cvpr}              


\usepackage{amsmath,amsfonts}
\usepackage{algorithmic}
\usepackage{algorithm}
\usepackage{array}
\usepackage{textcomp}

\usepackage{url}
\usepackage{verbatim}
\usepackage{graphicx}
\usepackage{svg}
\usepackage{multirow}
\usepackage{subcaption}
\usepackage{makecell}
\usepackage{stfloats}

\definecolor{cvprblue}{rgb}{0.21,0.49,0.74}
\usepackage[pagebackref,breaklinks,colorlinks,allcolors=cvprblue]{hyperref}

\def\paperID{XXXX} 
\def\confName{CVPR}
\def\confYear{2026}

\title{BEV-SLD: Self-Supervised Scene Landmark Detection for Global Localization with LiDAR Bird’s-Eye View Images}

\author{
David Skuddis \quad
Vincent Ress \quad
Wei Zhang \quad
Vincent Ofosu Nyako \quad
Norbert Haala\\
\small Institute for Photogrammetry and Geoinformatics, University of Stuttgart, Germany\\
{\tt\small
\{firstname.lastname, vincent.ofosu-nyako\}@ifp.uni-stuttgart.de
}
}

\begin{document}
\twocolumn[{%
  \renewcommand\twocolumn[1][]{#1}%
  \maketitle
  \begin{center}
    \centering
    \captionsetup{type=figure}
    \includegraphics[width=.93\textwidth]{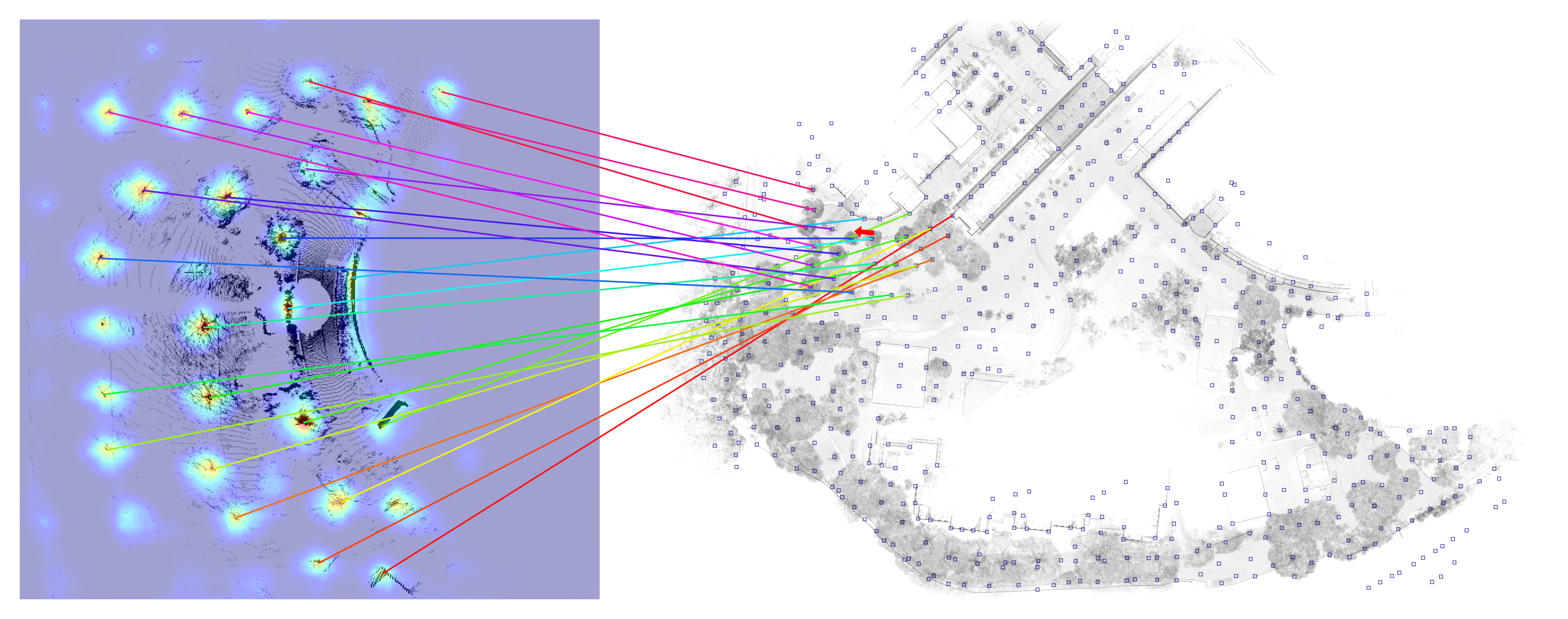}
    \captionof{figure}{Scene landmark-based localization on the MCD dataset~\cite{mcd_dataset}. \textbf{Left}: BEV density image from a local point cloud with predicted landmark locations. \textbf{Right}: Scene landmarks shown as blue squares. Lines indicate correctly predicted correspondences, and the red arrow marks the estimated pose. Real-time localization requires only a 20\,MB representation consisting of the network and landmark list.}
    \label{fig:teaser}
  \end{center}%
}]
\begin{abstract}
We present BEV-SLD, a LiDAR global localization method building on the Scene Landmark Detection (SLD) concept. Unlike scene-agnostic pipelines, our self-supervised approach leverages bird’s-eye-view (BEV) images to discover scene-specific patterns at a prescribed spatial density and treat them as landmarks. A consistency loss aligns learnable global landmark coordinates with per-frame heatmaps, yielding consistent landmark detections across the scene. Across campus, industrial, and forest environments, BEV-SLD delivers robust localization and achieves strong performance compared to state-of-the-art methods.
\end{abstract}

\section{Introduction}
LiDAR global localization remains challenging under low overlap, domain shifts, and at scales where dense maps are impractical. Many existing pipelines are scene-agnostic: retrieval methods encode each keyframe as a single feature vector, effective at high overlap but brittle under low overlap or domain shift, while keypoint-based approaches miss environment-specific regularities. Scene Coordinate Regression links local observations to global coordinates but requires learning too many locations and is sensitive to noise, which reduces accuracy. Inspired by vision-based Scene Landmark Detection (SLD) \cite{do2022learning,do2024improved} that uses SfM points as landmarks, we propose BEV-SLD\footnote{Code: \url{https://github.com/davidskdds/BEV-SLD}.} for LiDAR BEV images. In order to find suitable landmarks at an appropriate spatial density in any type of environment, we propose an approach for joint learning of landmark positions and detection. Our method learns scene-specific salient areas as landmarks via a new self-supervised scheme and links them to global coordinates through a compact landmark list. Inference requires no dense map, only a small network and a landmark list. 
Across four real-world datasets, BEV-SLD achieves state-of-the-art success rates, with the largest gains for queries far from the reference trajectory.

Our main contributions are:
\begin{itemize}
\item A new self-supervised, scene-specific landmark detector that learns recurring, coverage-aware landmarks across a reference sequence.
\item A scalable output design that couples high-resolution position heatmaps with low-resolution correspondence maps.
\item State-of-the-art success rates on real-world datasets using only a 20\,MB representation at inference, without a dense map.
\end{itemize}
\section{Related Work}
In LiDAR-based global localization, the objective is to estimate the pose of a query point cloud relative to a given map. The map typically consists of a sequence of keyframes with known poses in the global coordinate frame. Alternatively, a consolidated point cloud representing the entire environment may be used as the reference map.
While an obvious option is to use conventional global point cloud registration methods such as \cite{5152473,6751291,lim2024kiss}, these tend to require high computational resources with increasing map size, have difficulties with low overlaps, and are not optimized for scene-specific details that are beneficial for global localization.
In general, global localization methods can be divided into direct methods and retrieval-based methods. Direct methods are characterized by the direct estimation of a global pose, whereas retrieval-based methods usually compare the similarity between a query point cloud and a database to identify the nearest keyframe or submap. Retrieval-based methods address the problem commonly known as place recognition, but require additional local alignment to enable accurate global localization.
Both modern retrieval-based and direct methods still use a wide variety of point cloud representations as input, such as spherical range images \cite{10203089,10657300}, BEV images \cite{luo2024bevplaceplusplus,bonnbev} and raw (mostly downsampled) point clouds \cite{sgloc,Yang_2024_CVPR,cattaneo2022tro}.

\smallskip

\noindent\textbf{Retrieval-based Methods.}
A popular non-learning-based method for identifying similar keyframes is Scan Context and its variants \cite{8593953,9610172}, in which maximum point heights in a local polar grid serve as a descriptor. While this method and other rule-based methods such as BoW3D \cite{bow3d} and BTC \cite{btc} usually require less integration effort and have lower hardware requirements for the target platform, such as not requiring a GPU, their performance is usually inferior to learning-based methods. Examples of early learning-based retrieval-based approaches include PointNetVLAD \cite{8578568} and OverlapNet \cite{chen2020rss}.
Representative state-of-the-art methods include Locus \cite{vid2021locus}, OverlapTransformer \cite{ma2022ral}, MinkLoc3d \cite{minkloc_3d}, LoGG3D-Net \cite{vid2022logg3d}, EgoNN \cite{9645340}, BEVPlace++ \cite{luo2024bevplaceplusplus}, and LCDNet \cite{cattaneo2022tro}. The latter three additionally estimate relative poses and can therefore also be considered direct methods.

\smallskip

\noindent\textbf{Direct Methods.}
In addition to conventional methods for global point cloud registration such as FPFH \cite{5152473}, Go-ICP \cite{6751291}, 3D-BBS \cite{10610810} and KISS-Matcher \cite{lim2024kiss}, direct methods also include absolute pose regression (APR) methods such as HypLiLoc \cite{10203089}, PosePN++ \cite{YU2022108685} and DiffLoc \cite{10657300}. Here, a global pose is regressed directly by a neural network that receives a query point cloud as input.
Another class of direct methods is Scene Coordinate Regression (SCR), where global 3D coordinates are estimated for each point or pixel. These correspondences can subsequently be used to estimate a global pose. Representative state-of-the-art methods here include SGLoc \cite{sgloc}, LiSA \cite{Yang_2024_CVPR}, which integrates semantic awareness to improve robustness, and LightLoc \cite{li2025lightloc}.
One class of methods that has not yet been explored in the LiDAR domain is Scene Landmark Detection (SLD) \cite{do2022learning,do2024improved}, which has demonstrated particular accuracy and robustness in the vision domain compared to Scene Coordinate Regression (SCR) approaches.
A distinguishing feature of APR, SCR, and SLD compared to other approaches is that they do not require a map in the form of keyframes or a global point cloud. Instead, localization relies solely on neural networks, which typically have a smaller memory footprint than map-based methods.
\begin{figure*}[t]
\centering
\includesvg[width=0.99\textwidth]{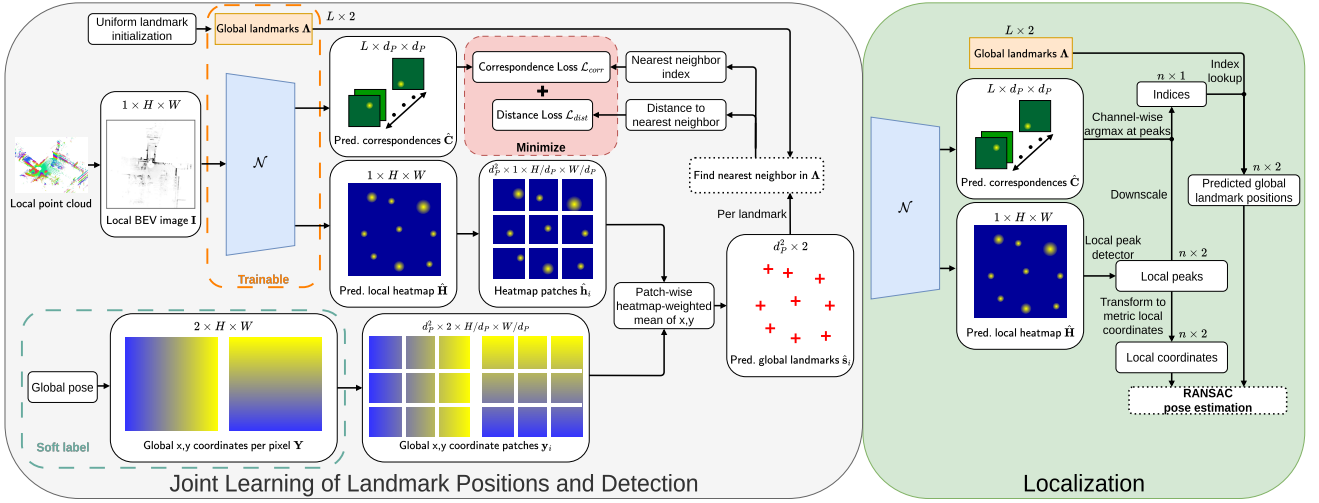} 
\caption{Overview of the proposed landmark learning framework (left) and the localization pipeline at inference time (right).}
\label{fig:overview}
\end{figure*}
\section{Methodology}
We propose a method to estimate the global 3 DoF pose of a local 3D LiDAR point cloud within a pre-mapped environment by detecting so-called scene landmarks. Scene landmarks are scene-specific locations distributed across the map that serve as localization anchors.

Our approach assumes a reference sequence of the environment consisting of local point clouds together with their associated global poses. These global poses can be obtained using LiDAR SLAM methods such as \cite{dellenbach2021cticp,10610818}. For reliable localization, the reference sequence should provide dense coverage of the target environment. In the following, coordinates defined in the scene or map are referred to as global, while coordinates defined in the sensor frame are referred to as local.

During training (see Section \ref{sec:selfsupervised}), the local point clouds of the reference sequence and their corresponding global poses are used to jointly learn scene-specific landmark locations and their detection. To enable efficient training, local point clouds are represented as bird’s-eye-view (BEV) density images obtained by projecting LiDAR points onto the local $xy$-plane (see Section \ref{sec:preproc}). Compared to spherical range images, BEV representations avoid scaling effects caused by viewpoint changes and are computationally more efficient than operating directly on 3D point clouds.

In Scene Landmark Detection (SLD), the network must predict not only the locations of landmarks in the current observation but also their assignment to a predefined landmark set with global coordinates. This leads to a trade-off in existing SLD methods \cite{do2022learning,do2024improved} between spatial resolution, which is important for accurate landmark localization, and scalability. If a high spatial resolution $H \times W$ is combined with a large number of landmarks $L$, the network output becomes memory intensive, as the number of predicted values grows to $L \times H \times W$.
To decouple accuracy and scalability we propose a breakdown of the problem into two sub-problems:
\begin{enumerate}
  \item Landmark detection: Detect scene landmarks in the current observation as accurately as possible using high spatial resolution and a single output channel.
  \item Correspondence prediction: Predict a pixel-wise probability distribution over scene landmarks for each region in the BEV image using lower spatial resolution and a larger channel dimension.
\end{enumerate}

\smallskip

\noindent\textbf{Overview.}
An overview of the method is provided in Fig.~\ref{fig:overview}.
Our pipeline comprises three main stages: preprocessing (Section \ref{sec:preproc}), landmark learning (Section \ref{sec:selfsupervised}), and localization (Section \ref{sec::loc}). The stages will be presented in detail below.
\subsection{Preprocessing}\label{sec:preproc}
The following sections detail the preprocessing steps used in our approach.

\smallskip

\noindent\textbf{Keyframe Selection.}
Keyframes of the reference sequences are chosen based on a distance threshold that enforces a minimum spatial separation between consecutive frames.
This threshold is dataset-dependent and ranges from 0.3\,m to 0.5\,m.

\smallskip

\noindent\textbf{Bird’s-Eye View Density Image Generation.}
We follow the work in \cite{luo2024bevplaceplusplus} and \cite{bonnbev} to create LiDAR Bird’s-Eye View (BEV) density images.
A LiDAR point cloud, which can be either a single scan or a submap, is first downsampled using a voxel grid filter.
All points are projected into the local $xy$-plane and divided into 2D grid cells. For each grid cell the number of points is counted. As normalization, every pixel is then divided by the maximum number of points within a pixel.
For all experiments presented, images of 512 $\times$ 512 pixels with a pixel size of 0.2\,m are used. The images thus represent an area of 102.4\,m $\times$ 102.4\,m. Experiments have shown that these parameters are suitable for all environments and sensor configurations tested. For narrower indoor scenarios or LiDAR scanners with a shorter range, parameter changes for a smaller field of view but higher-resolution BEV images may be advantageous.

\smallskip

\noindent\textbf{Data Augmentation.}
To improve robustness to positional and orientational variations, random translations and rotations are applied to the BEV images during training. Translations range from \(-25\%\) to \(25\%\) of the image dimensions, corresponding to shifts of up to approx. $\pm25$\,m. Rotations are sampled uniformly from the full range of \(0^\circ\) to \(360^\circ\).
Furthermore, to handle point density differences caused by varying recording positions, the entire BEV image is scaled by a random factor between 0.5 and 1.5.

\smallskip

\noindent\textbf{Landmark Initialization.}
Initially, the scene is covered with landmarks, positioned on the basis of the global poses of the reference sequence. Here, for every global keyframe pose of the reference sequence $d_P^2$ landmarks are generated in the global area covered by the respective BEV density image. In detail, the covered area is divided into $d_P \times d_P$ patches and in every patch center a landmark is sampled.
Since we generate $d_P^2$ landmarks per keyframe, which leads to a large number of overlapping landmarks, we apply a grid based downsampling to obtain the initial global landmark list $\mathbf{\Lambda}$. 
The grid filter size $s_{grid}$ and the number of patches $d_P^2$ control the landmark density (see supplementary material for details).
\subsection{Joint Learning of Landmark Positions and Detection}\label{sec:selfsupervised}
\begin{figure}[]
  \centering
  \includesvg[width=0.49 \textwidth]{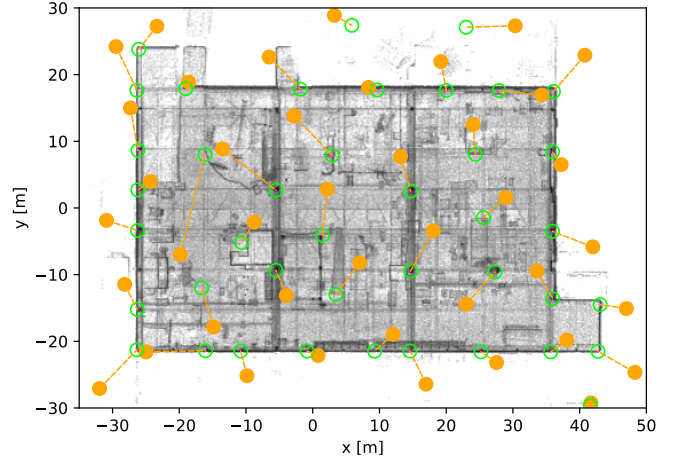}
  \caption{Initial landmarks and landmarks after joint learning of landmark positions and detection in a factory floor environment: filled \textbf{orange} circles represent landmark positions at initialization and \textbf{green} non-filled circles after training. Orange lines connect corresponding landmarks. The background point cloud is shown for visualization only.}
  \label{fig:landmark_movement}
\end{figure}

To identify suitable landmarks at an appropriate spatial density in diverse environments, we propose a self-supervised approach that jointly learns landmark positions and their detection.
The proposed framework is designed with three objectives: (i) \textbf{landmark refinement} -- adjust the landmark positions $\mathbf{\Lambda}$ toward salient and geometrically stable locations (as shown in Fig.~\ref{fig:landmark_movement}); (ii) \textbf{precise detection} -- enable precise localization of these refined landmarks through heatmap prediction; (iii) \textbf{index estimation} -- determine the correspondence between each detected local landmark and its entry in the global landmark set.
This is achieved by enforcing a consistency loss between the predicted heatmap $\hat{\mathbf{H}}$ and the global landmark set $\mathbf{\Lambda}$, complemented by an additional loss term for correspondence estimation.
In detail, this is realized using a network $\mathcal{N}$ with trainable parameters $\theta$, which takes a BEV density image $\mathbf{I} \in \mathbb{R}^{1 \times H \times W}$ as input and predicts both a local landmark heatmap $\hat{\mathbf{H}} \in \mathbb{R}^{1 \times H \times W}$ and a correspondence map $\hat{\mathbf{C}} \in \mathbb{R}^{L \times d_P \times d_P}$, where $L$ denotes the number of global landmarks.
\begin{equation}
(\hat{\mathbf{H}}, \hat{\mathbf{C}}) = \mathcal{N}(\mathbf{I}; \theta)
\end{equation}
The crucial part now is to link local heatmap predictions $\hat{\mathbf{H}}$ with the global landmark set $\mathbf{\Lambda}$. We solve this by creating a coordinate map $\mathbf{Y} \in \mathbb{R}^{2 \times H \times W}$ for each local BEV image, which provides the x- and y-coordinates in the global frame for each pixel. The heatmaps are then used as a weighting factor and combined with the coordinate maps to obtain differentiable global landmark estimates. In order to detect more than just one landmark, we divide predicted heatmaps $\hat{\mathbf{H}}$ and the corresponding coordinate maps $\mathbf{Y}$ into $d_P \times d_P$ patches, which we call $\hat{\mathbf{h}}_i\in \mathbb{R}^{1 \times \frac{H}{d_P} \times \frac{W}{d_P}}$ and $\mathbf{y}_i\in \mathbb{R}^{2 \times \frac{H}{d_P} \times \frac{W}{d_P}}$ respectively.
To obtain predicted global coordinates $\hat{\mathbf{s}}_i$ for each patch, we apply softmax to $\hat{\mathbf{h}}_i$ and then form a weighted sum of all coordinates $\mathbf{y}_i$ of the patch:
\begin{equation}
\hat{\mathbf{s}}_i(\hat{\mathbf{h}}_i,\mathbf{y}_i) = 
\begin{bmatrix}
x_i \\
y_i
\end{bmatrix}
=
\sum_{u,v} \mathrm{softmax}(\hat{\mathbf{h}}_i)_{u,v}
\begin{bmatrix}
(\mathbf{y}_i)_{0,u,v} \\
(\mathbf{y}_i)_{1,u,v}
\end{bmatrix}
\end{equation}
The predicted global coordinates $\hat{\mathbf{s}}_i$ per patch are now utilized in two different loss terms, namely a distance loss $\mathcal{L}_{dist}$ and a correspondence loss $\mathcal{L}_{corr}$, which together form the resulting loss function $\mathcal{L}_{total}$.
\begin{equation}
\mathcal{L}_{total} = \alpha \, \mathcal{L}_{dist}+\beta \,\mathcal{L}_{corr}
\end{equation}
$\alpha$ and $\beta$ are weight parameters to balance the impact of each term.

\smallskip

\noindent\textbf{Distance Loss.}
The distance loss $\mathcal{L}_{dist}$ penalizes the deviations between the predicted global coordinates $\hat{\mathbf{s}}_i$ and their respective nearest neighbors in $\mathbf{\Lambda}$.
A logarithmic scaling is applied to attenuate the influence of large deviations, and a scaling parameter $\gamma$ controls the effect.
\begin{equation}
\mathcal{L}_{dist}(\hat{\mathbf{H}}, \mathbf{\Lambda},\mathbf{Y}) = \sum_{i=1}^{d_P^2} \log\left(1 +\gamma \min_{j = 1, \ldots, L} \left\| \hat{\mathbf{s}}_i- \mathbf{\Lambda}_j \right\|_2\right)
\end{equation}
This loss term, which is independent of the predicted correspondences $\hat{\mathbf{C}}$, enforces consistency and causes the global set of coordinates $\mathbf{\Lambda}$, which is also learnable, to move towards distinctive locations.

\smallskip

\noindent\textbf{Correspondence Loss.}
The predicted correspondence tensor 
$\hat{\mathbf{C}} \in \mathbb{R}^{L \times d_P \times d_P}$ 
is defined to share the spatial resolution of the patch grid $d_P \times d_P$. 
Thus, each patch, and consequently each predicted global landmark position 
$\hat{\mathbf{s}}_i$, is associated with a correspondence vector 
$\hat{\mathbf{c}}_i$. 
For supervision, the label of $\hat{\mathbf{c}}_i$ is given by the index 
$j^{*}_i$ of the nearest landmark in $\mathbf{\Lambda}$ to $\hat{\mathbf{s}}_i$, 
\begin{equation}
j^{*}_i = \underset{j = 1, \ldots, L}{\arg\min} \; 
\left\| \hat{\mathbf{s}}_i - \mathbf{\Lambda}_j \right\|_2 .
\end{equation}
The correspondence loss $\mathcal{L}_{\mathrm{corr}}$ enforces agreement 
between $\hat{\mathbf{C}}$ and the nearest-neighbor indices by applying a 
softmax to each $\hat{\mathbf{c}}_i$ and computing the categorical 
cross-entropy with the one-hot label of $j^{*}_i$.

\smallskip

\noindent\textbf{Training Process.}
The process can be described with the following formula:
\begin{equation}
\min_{\theta, \mathbf{\Lambda}} \; \mathcal{L}_{total}(\mathcal{N}(\mathbf{I}; \theta), \mathbf{\Lambda},\mathbf{Y})
\end{equation}
During training, the network parameters $\theta$ are optimized to predict the landmark heatmaps, while the landmark coordinates $\mathbf{\Lambda}$ are jointly refined. For practical reasons, we represent the global scene landmark coordinates $\mathbf{\Lambda}$ as a differentiable embedding within the network $\mathcal{N}$ in our implementation. Notably, the global landmarks are input-independent and remain constant across different inputs.
Patches that contain no LiDAR points are excluded during training. If the distance between $\hat{\mathbf{s}}_i$ and the nearest landmark in $\mathbf{\Lambda}$ is greater than half the patch diagonal, they are not taken into account in the correspondence loss $\mathcal{L}_{\mathrm{corr}}$. For clarity of the formulation, these filtering steps are omitted from the above equations. The patch generation and resulting loss function are outlined in Fig.~\ref{fig:overview}.
At the beginning, neither meaningful local landmarks are predicted by the heatmaps $\hat{\mathbf{H}}$, nor are the scene landmarks $\mathbf{\Lambda}$ placed in advantageous locations. This can be seen as a circular problem in which local landmarks that are advantageous to predict can trigger a shift in the scene landmarks $\mathbf{\Lambda}$ and vice versa.
Nevertheless, we observe that landmarks converge toward distinctive locations such as building corners or tree trunks and the network learns to highlight these salient areas in the predicted landmark heatmap.
We hypothesize that landmarks near salient structures are easier to predict, enabling the network to associate similar features elsewhere and gradually attract remaining landmarks toward distinctive locations.

\smallskip

\noindent\textbf{Network Architecture.}\label{sec::method_arch}
Using BEV density images allows us to leverage image semantic segmentation architectures, which provide dense predictions with global context. To meet our requirements (i) compatibility with resource-constrained platforms, (ii) lightweight design for efficient training, (iii) minimum inference speed of 10 fps, and (iv) the ability to capture global context, we adopt a modified Feature Pyramid Network \cite{seferbekov2018feature}, which we empirically found to be particularly well suited for the task. Specifically, we reduce feature channels, remove intermediate upsampling for efficiency, add downsampling blocks to better capture context, and introduce a second output branch for correspondence prediction. Figure \ref{fig:network_arch} illustrates the resulting network architecture. All blocks use layer normalization and LeakyReLU as activation. The final upsampling layer in the heatmap prediction branch employs bicubic interpolation to enhance spatial precision for landmark prediction, while all preceding upsampling layers use bilinear interpolation for computational efficiency. Overall, the network comprises approximately 4.7 million parameters.
\begin{figure}[h]
\centering
\includesvg[width=0.49 \textwidth]{fig/det_class_net.svg}
\caption{Network architecture. \textbf{Res. Blocks} represent residual blocks \cite{res_block}, \textbf{Down Blocks} consist of max pooling followed by a residual block and \textbf{Conv. Blocks} consist of convolution followed by layer normalization and activation function.}
\label{fig:network_arch}
\end{figure}
\subsection{Localization} \label{sec::loc}
For localization, we assume that a network has already been trained for a specific environment and is now used to localize new point clouds in the same scene.
The BEV density image generated from the current point cloud is first fed into the network to obtain predicted landmark heatmap and correspondence map. We then apply the local peak detector (\textit{peak\_local\_max}) of \cite{van2014scikit} to the predicted heatmap to detect the strongest $n$ local maxima. For each maximum, based on the correspondence output $\hat{\mathbf{C}}$ at the corresponding pixel position, the channel index with the highest correspondence score is determined. This index is then used to extract the corresponding global landmark position using the learned landmark coordinates $\mathbf{\Lambda}$.
By converting the local pixel coordinates of the local maxima into local metric $xy$-coordinates, pairs of local and global landmark positions are obtained. These are used in the final step to estimate a global pose ($x$, $y$ and azimuth) using RANSAC \cite{fischler1981random}. An overview of the localization pipeline is shown in Fig. \ref{fig:overview}.
\section{Experiments \& Results}
\begin{figure*}[htp]
\centering
\includegraphics[width=0.91\textwidth]{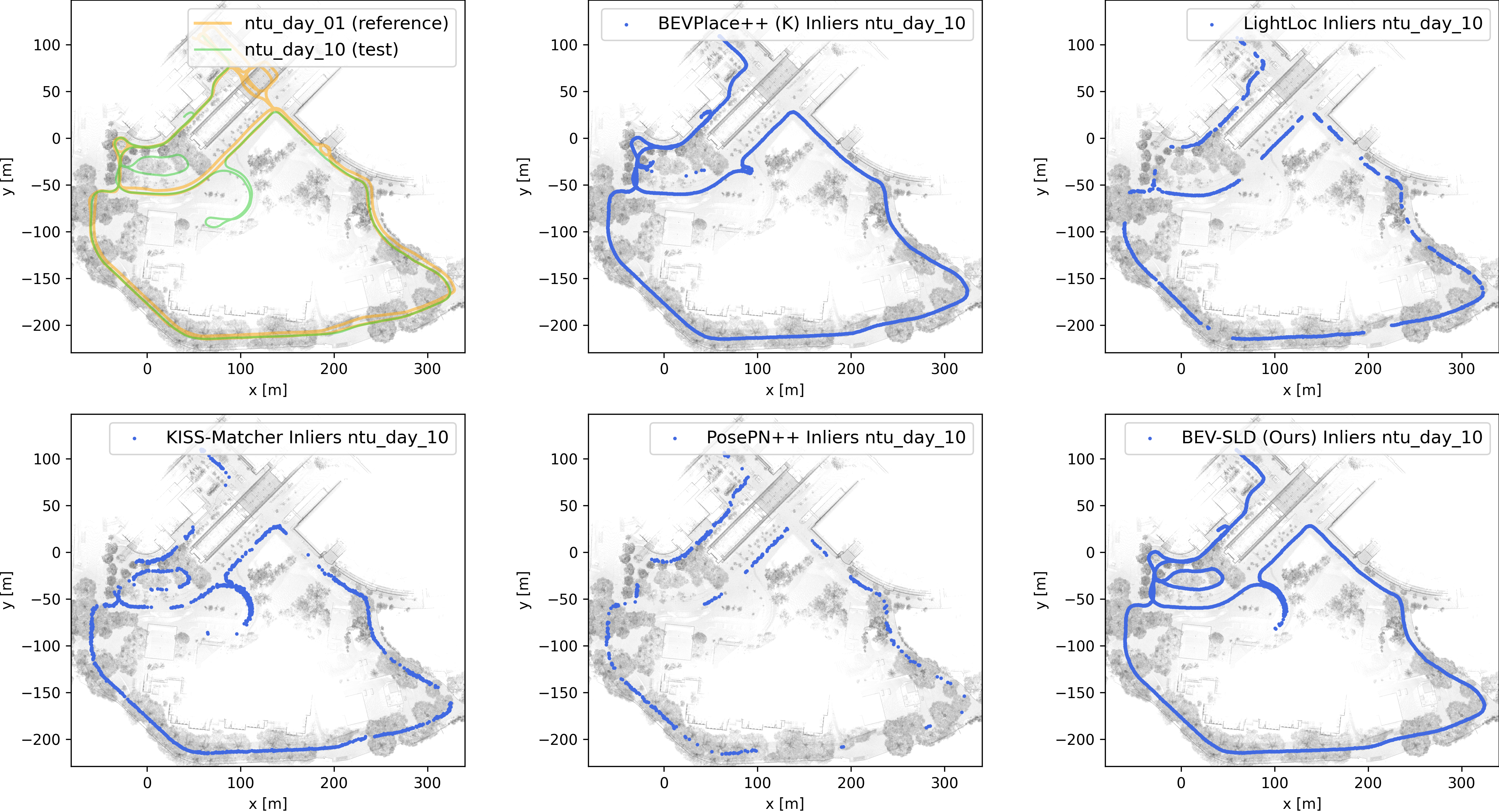}
\caption{Qualitative comparison of BEVPlace++ \cite{luo2024bevplaceplusplus} trained on KITTI (K), LightLoc \cite{li2025lightloc}, KISS-Matcher \cite{lim2024kiss}, PosePN++ \cite{YU2022108685} and the proposed method. 
\textbf{Top left}: reference trajectory (ntu\_day\_01, orange) and ground truth trajectory of the test sequence (ntu\_day\_10, green).
\textbf{Others}: inliers (errors $<$2\,m, 5$^\circ$) of the different methods. 
Our method achieves broader inlier pose coverage than the baselines.
}
\label{fig:day_10_hor_plot}
\end{figure*}
We first describe the datasets and metrics, then present ablations, followed by comparisons to state-of-the-art global localization methods.
\subsection{Datasets}\label{sec:datasets}
For our experiments, we use three public datasets and one self-collected dataset.
We conducted our experiments on sequences from the MCD dataset \cite{mcd_dataset}, which captures an urban campus area including a park using an Ouster OS1-128 sensor; on the NCLT dataset \cite{nclt_dataset}, which represents a campus environment recorded with a Velodyne HDL-32E sensor; on forest sequences from the Wild-Places dataset \cite{wild_places} recorded by a Velodyne VLP-16 LiDAR; and on self-recorded sequences collected in an industrial factory floor setting using an Ouster OS1-128 sensor. While single scans are used for training and localization on the MCD dataset, the NCLT dataset, and our self-recorded dataset, the Wild-Places dataset provides submaps that accumulate 0.5 seconds of LiDAR point clouds. Table \ref{tab:datasets} provides an overview of the datasets.
\begin{table}[h]
\centering
\caption{Overview of datasets. Reported are \# Seq., accumulated distance (Acc. Dist.), acquisitions time span, bounding box area, and environment type.}
\label{tab:datasets}
\resizebox{0.9\columnwidth}{!}{%
\begin{tabular}{|l|llll|}
\hline
Name & MCD \cite{mcd_dataset} & NCLT \cite{nclt_dataset} & Wild-Places \cite{wild_places} & Factory Floor \\ \hline
\# Seq. & 6 & 8 & 4 & 2 \\
Acc. Dist. & 10.7\,km & 43.3\,km & 12.7\,km & 875\,m \\
Time Span & 13\,days & 4\,months & 14\,months & 9\,months \\
Area [m$\times$m] & $550\times400$ & $700\times400$ & $800\times600$ & $70\times40$ \\
Environment & Campus & Campus & Forest & Industrial \\ \hline
\end{tabular}
}
\end{table}
\subsection{Resource Usage \& Runtime Analysis}
Experiments were conducted on a workstation featuring an AMD Ryzen Threadripper PRO 5955WX (16 cores), 256\,GB RAM, and an NVIDIA RTX 6000 Ada GPU with 48\,GB VRAM.

\smallskip

\noindent\textbf{Training Time.}
The training time depends heavily on the size of the reference sequence. Table \ref{tab:training_time} provides an overview of the number of epochs and training time in our evaluation.
\begin{table}[h]
\centering
\caption{Training time per reference sequence.}
\label{tab:training_time}
\resizebox{0.9\columnwidth}{!}{%
\begin{tabular}{|l|llll|}
\hline
Dataset & Seq. & \# Frames & \# Epochs & Duration {[}h{]} \\ \hline
MCD & ntu\_day\_01 & 5,694 & 600 & 14 \\
NCLT & 2012-01-15 & 15,026 & 300 & 12 \\
Wild-Places & V-03 & 6,757 & 800 & 16.5 \\
Factory Floor & 2024-01 & 1,031 & 50 & 1 \\ \hline
\end{tabular}
}
\end{table}

\smallskip

\noindent\textbf{Localization Runtime and Memory Footprint.}
The localization time remains approximately constant across all sequences. First the network predicts a landmark heatmap and correspondences based on a BEV input image, then the local peak detector extracts landmarks from the predicted heatmap, which are then used to estimate the pose via RANSAC using the predicted correspondences from the network, resulting in an overall speed of approximately 35 FPS.
The network parameters and landmarks require approximately 20\,MB of memory.
\subsection{Evaluation Metrics}\label{sec:metrics}
Two metrics are used to evaluate global localization performance.
As used in \cite{10494918,luo2024bevplaceplusplus}, a success rate (SR) is calculated, which shows the percentage of poses with errors smaller than 2\,m and $5^\circ$ compared to ground truth. Since reference sequences may not cover all test areas, SR is meaningful only for relative comparison. In addition, we calculate median translational error TE and median rotational error RE. We decided not to report the mean absolute errors, as these are often not meaningful due to the influence of outliers. Since the success rates achieved in most sequences are higher than 50\,\%, the medians can be considered a more meaningful indicator of accuracy.
\subsection{Experimental Analysis and Ablation Studies}
In this section, we analyze the proposed approach through several experiments.

\smallskip

\noindent\textbf{Frozen Initial Landmarks vs. Self-Supervised Learned Landmarks.}\label{sec:ss_lms}
In this experiment, we froze the initial landmarks and trained the network without optimizing the landmark positions $\Lambda$.
We then compare the localization results with the proposed variant, in which the landmark positions are also optimized during training (see Section~\ref{sec:selfsupervised}).
We trained and evaluated on sequences from the MCD dataset \cite{mcd_dataset} and used the same configuration parameters and trained the same number of epochs in the experiments.
Table~\ref{tab:abl_ss} presents the results of the experiments. The version with optimized landmark positions clearly outperforms the one with frozen landmarks, achieving a higher localization success rate as well as significantly lower translational and rotational errors.
\begin{table}[h]
\centering
\caption{Localization results using frozen initial and optimized landmarks, evaluated on MCD dataset scenes.}
\label{tab:abl_ss}
\resizebox{0.9\columnwidth}{!}{%
\begin{tabular}{|l|lll|lll|}
\hline 
\multirow{2}{*}{\shortstack{Seq. ntu\_x\\(reference: day\_01)}} & \multicolumn{3}{l|}{Frozen Initial Landmarks} & \multicolumn{3}{l|}{Optimized Landmarks} \\ \cline{2-7} 
 & \multicolumn{1}{l|}{\begin{tabular}[c]{@{}l@{}}SR ↑\\ {[}\%{]}\end{tabular}} & \multicolumn{1}{l|}{\begin{tabular}[c]{@{}l@{}}TE ↓\\{[}m{]}\end{tabular}} & \begin{tabular}[c]{@{}l@{}}RE ↓\\{[}deg{]}\end{tabular} & \multicolumn{1}{l|}{\begin{tabular}[c]{@{}l@{}}SR ↑\\ {[}\%{]}\end{tabular}} & \multicolumn{1}{l|}{\begin{tabular}[c]{@{}l@{}}TE ↓\\{[}m{]}\end{tabular}} & \begin{tabular}[c]{@{}l@{}}RE ↓\\{[}deg{]}\end{tabular} \\ \hline
day\_02 & 68.05 & 1.09 & 1.63 &\textbf{100.00} & \textbf{0.19} & \textbf{0.27} \\ 
day\_10 &40.35 & 2.43 & 4.00 &\textbf{92.04} & \textbf{0.20} & \textbf{0.44} \\ 
night\_04 &38.65 & 2.59 & 4.06 &\textbf{91.82} & \textbf{0.21} & \textbf{0.44} \\ 
night\_08 &48.39 & 1.88 & 2.93 &\textbf{95.26} & \textbf{0.21} & \textbf{0.47} \\ 
night\_13 &53.64 & 1.64 & 2.39 &\textbf{99.91} & \textbf{0.20} & \textbf{0.39} \\ \hline
\end{tabular}%
}
\end{table}

\smallskip

\noindent\textbf{Qualitative Landmark Analysis.}\label{sec:qual_analysis}
To gain an impression of which locations the system selects as landmarks, in Fig. \ref{fig:landmark_movement} we have illustrated the initial and learned landmarks for the Factory Floor scene. It is noticeable that during training, some landmarks move to corner points and other distinctive locations that may be easy to detect. This reflects exactly the desired behavior: without explicitly defining corner points, these are selected as suitable landmarks during training. Some landmarks at the image boundaries exhibit limited displacement, as they are observed in only a few keyframes, resulting in low contribution to the loss.

\smallskip

\noindent\textbf{Scalability to Larger Scenes.}
To evaluate performance on larger real-world scenes, we trained 800\,ep.\,/\,23\,h on MulRan’s \cite{gskim-2020-mulran} 23.4\,km Sejong City 2019-06-20 sequence and tested on 2019-08-20. 78.0\,\% of poses achieve errors $<$7\,m\,/\,7° and 64.1\% errors $<$2\,m\,/\,5°, using 10,769 landmarks.
To further assess scalability with respect to landmark count, we evaluated an untrained model initialized with 100k landmarks and compared it to our MCD model (614 landmarks): network runtime increases marginally from 17.3\,ms to 17.8\,ms, while parameters grow from 4.96\,M to 43.4\,M and memory footprint from 20\,MB to 173.7\,MB. 

\smallskip

\begin{figure}[t]
\centering
\includesvg[width=0.49 \textwidth]{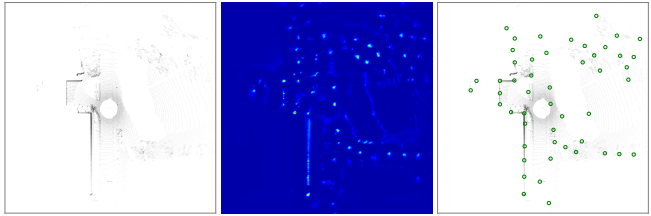}
\caption{Heatmap prediction for an unseen scene. \textbf{Left}: MCD dataset BEV image. \textbf{Mid}: Heatmap prediction using a Factory Floor dataset trained network. \textbf{Right}: Detected local peaks.}
\label{fig:unseen_envs}
\end{figure}

\noindent\textbf{Applicability to Unseen Environments.}
Our training and localization pipeline is inherently scene-specific: the network is trained for a particular environment, and predicted correspondences are tied to a dedicated global landmark set. Consequently, the full approach is not directly applicable to unseen environments.
However, to assess potential transferability of the heatmap predictor, we applied a network trained on our Factory Floor reference sequence to BEV images from the MCD dataset. We limit this evaluation to a qualitative analysis. The results are promising: geometrically distinctive structures are highlighted as landmark candidates, and corners, for example, are detected reliably across consecutive frames. Fig. \ref{fig:unseen_envs} (best viewed when zoomed in) shows a representative example.
\begin{figure}[b]
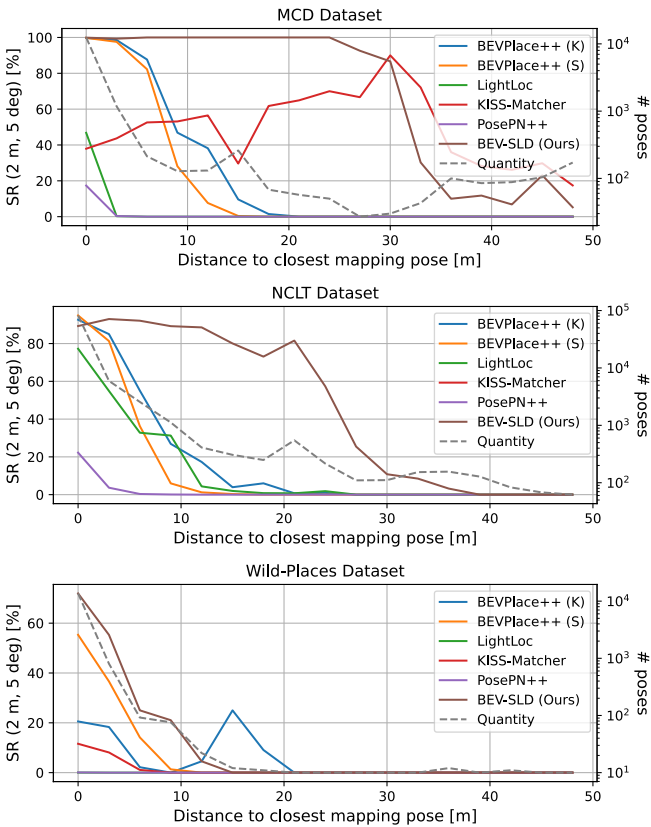

  \centering
  \includesvg[width=0.49\textwidth]{fig/succ_over_dist_mcd_posepn.svg} 
  
  \vspace{0.2cm}
  \includesvg[width=0.49\textwidth]{fig/succ_over_dist_nclt_posepn.svg} 

\vspace{0.2cm}
  \includesvg[width=0.49\textwidth]{fig/succ_over_dist_wild_places_posepn.svg} 
  \caption{Success rate as a function of the distance to the closest reference pose, 
aggregated over all sequences. Distances are grouped into 3\,m bins.}
  \label{fig:sr_vs_distance}
\end{figure}
\begin{table*}[ht]
\centering
\caption{Quantitative results across datasets. We report median translational error (TE), median rotational error (RE), and success rate SR (errors $<$2\,m, 5$^\circ$). Best results are highlighted in \textbf{bold}.}
\label{tab:results}
\resizebox{0.99\textwidth}{!}{%
\begin{tabular}{|l|l|lll|lll|lll|lll|lll|lll|}
\cline{1-20}
\multirow{2}{*}{ \shortstack{Dataset \\ (reference sequence)} } 
& \multirow{2}{*}{Seq.} 
& \multicolumn{3}{l|}{BEVPlace++ (K) \cite{luo2024bevplaceplusplus}} 
& \multicolumn{3}{l|}{BEVPlace++ (S)  \cite{luo2024bevplaceplusplus}} 
& \multicolumn{3}{l|}{LightLoc \cite{li2025lightloc}} 
& \multicolumn{3}{l|}{KISS-Matcher \cite{lim2024kiss}} 
& \multicolumn{3}{l|}{PosePN++ \cite{YU2022108685}} 
& \multicolumn{3}{l|}{\begin{tabular}[c]{@{}l@{}}BEV-SLD (Ours)\end{tabular}} \\ \cline{3-20}
&  
& \multicolumn{1}{l|}{\begin{tabular}[c]{@{}l@{}}SR ↑\\ {[}\%{]}\end{tabular}} 
& \multicolumn{1}{l|}{\begin{tabular}[c]{@{}l@{}}TE ↓\\{[}m{]}\end{tabular}} 
& \begin{tabular}[c]{@{}l@{}}RE ↓\\{[}deg{]}\end{tabular} 
& \multicolumn{1}{l|}{\begin{tabular}[c]{@{}l@{}}SR ↑\\ {[}\%{]}\end{tabular}} 
& \multicolumn{1}{l|}{\begin{tabular}[c]{@{}l@{}}TE ↓\\{[}m{]}\end{tabular}} 
& \begin{tabular}[c]{@{}l@{}}RE ↓\\{[}deg{]}\end{tabular} 
& \multicolumn{1}{l|}{\begin{tabular}[c]{@{}l@{}}SR ↑\\ {[}\%{]}\end{tabular}} 
& \multicolumn{1}{l|}{\begin{tabular}[c]{@{}l@{}}TE ↓\\{[}m{]}\end{tabular}} 
& \begin{tabular}[c]{@{}l@{}}RE ↓\\{[}deg{]}\end{tabular} 
& \multicolumn{1}{l|}{\begin{tabular}[c]{@{}l@{}}SR ↑\\ {[}\%{]}\end{tabular}} 
& \multicolumn{1}{l|}{\begin{tabular}[c]{@{}l@{}}TE ↓\\{[}m{]}\end{tabular}} 
& \begin{tabular}[c]{@{}l@{}}RE ↓\\{[}deg{]}\end{tabular} 
& \multicolumn{1}{l|}{\begin{tabular}[c]{@{}l@{}}SR ↑\\ {[}\%{]}\end{tabular}} 
& \multicolumn{1}{l|}{\begin{tabular}[c]{@{}l@{}}TE ↓\\{[}m{]}\end{tabular}} 
& \begin{tabular}[c]{@{}l@{}}RE ↓\\{[}deg{]}\end{tabular} 
& \multicolumn{1}{l|}{\begin{tabular}[c]{@{}l@{}}SR ↑\\ {[}\%{]}\end{tabular}} 
& \multicolumn{1}{l|}{\begin{tabular}[c]{@{}l@{}}TE ↓\\{[}m{]}\end{tabular}} 
& \begin{tabular}[c]{@{}l@{}}RE ↓\\{[}deg{]}\end{tabular} \\ \cline{1-20}
MCD ntu\_x & day\_02 & \textbf{100.00} & 0.27 & 0.45 & 99.96 & 0.29 & 0.50 & 43.50 & 1.36 & 4.57 & 40.93 & 1.51 & 4.61 & 15.80 & 4.00 & 3.99 & \textbf{100.00} & \textbf{0.19} & \textbf{0.27} \\ \cline{2-2}
(day\_01) & day\_10 & 81.53 & 0.22 & 0.54 & 80.24 & 0.26 & 0.59 & 42.72 & 1.20 & 5.41 & 43.71 & 1.25 & 5.01 & 13.35 & 4.54 & 3.73 & \textbf{92.04} & \textbf{0.20} & \textbf{0.44} \\ \cline{2-2}
 & night\_04 & 86.74 & 0.46 & 0.60 & 86.13 & 0.48 & 0.62 & 14.27 & 14.68 & 6.23 & 14.78 & 14.53 & 6.29 & 9.71 & 29.21 & 12.94 & \textbf{91.82} & \textbf{0.21} & \textbf{0.44} \\ \cline{2-2}
 & night\_08 & 87.51 & 0.23 & 0.56 & 86.82 & 0.26 & 0.60 & 40.54 & 1.53 & 4.69 & 40.06 & 1.60 & 4.79 & 14.09 & 4.18 & 3.63& \textbf{95.26} & \textbf{0.21} & \textbf{0.47} \\ \cline{2-2}
 & night\_13 & 99.87 & \textbf{0.20} & 0.50 & 99.31 & 0.22 & 0.54 & 48.55 & 1.07 & 4.01 & 49.63 & 1.18 & 3.92 & 18.23 & 3.45 & 2.74 & \textbf{99.91} & \textbf{0.20} & \textbf{0.39} \\ \cline{1-20}
NCLT 2012-x & 01-22 & 87.91 & \textbf{0.38} & 1.32 & 88.15 & 0.44 & 1.38 & 66.71 & 0.70 & 1.78 &  & \textsc{n/a} &  & 22.87 & 4.22 & 3.38 & \textbf{88.57} & 0.43 & \textbf{1.13} \\ \cline{2-2}
(01-15) & 02-02 & 89.53 & \textbf{0.42} & 1.06 & 90.01 & 0.46 & 1.05 & 69.03 & 0.67 & 1.33 &  & \textsc{n/a} &  & 20.36 & 4.80 & 2.86 & \textbf{92.30} & \textbf{0.42} & \textbf{0.74} \\ \cline{2-2}
 & 02-18 & 91.27 & \textbf{0.30} & 0.98 & \textbf{92.31} & 0.36 & 1.01 & 83.74 & 0.54 & 0.99 &  & \textsc{n/a} &  & 28.00 & 3.00 & 1.99 & 91.26 & 0.40 & \textbf{0.81} \\ \cline{2-2}
 & 02-19 & \textbf{91.20} & \textbf{0.34} & 1.01 & 90.75 & 0.40 & 1.05 & 71.42 & 0.63 & 1.25 &  & \textsc{n/a} &  & 24.14 & 3.85 & 2.73 & 89.63 & 0.41 & \textbf{0.84} \\ \cline{2-2}
 & 03-31 & 91.50 & \textbf{0.31} & 1.05 & \textbf{91.95} & 0.37 & 1.12 & 80.14 & 0.64 & 1.14 &  & \textsc{n/a} &  & 22.32 & 3.80 & 2.30& 89.68 & 0.41 & \textbf{0.87} \\ \cline{2-2}
 & 05-11 & 79.34 & \textbf{0.48} & 1.37 & \textbf{82.27} & 0.53 & 1.38 & 60.95 & 1.10 & 2.01 &  & \textsc{n/a} &  & 8.86 & 13.20 & 5.56 & 82.10 & 0.52 & \textbf{1.05} \\ \cline{2-2}
 & 05-26 & 83.87 & \textbf{0.41} & 1.26 & 85.19 & 0.47 & 1.27 & 71.38 & 0.92 & 1.58 &  & \textsc{n/a} &  & 11.46 & 8.69 & 3.60 & \textbf{85.74} & 0.51 & \textbf{0.95} \\ \cline{1-20}
Wild-Places & V-01 & 27.50 & 72.72 & 17.41 & 69.01 & 1.46 & \textbf{0.75} & 3.24 & 16.77 & 15.81 & 14.77 & 200.54 & 54.17 & 0.17 & 42.62 & 41.74 & \textbf{82.08} & \textbf{1.07} & 0.99 \\ \cline{2-2}
(V-03) & V-02 & 16.75 & 138.80 & 44.56 & 42.45 & 2.21 & \textbf{1.04} & 2.73 & 43.07 & 46.21 & 13.17 & 220.08 & 61.72 & 0.02 & 46.47 & 122.90 & \textbf{63.96} & \textbf{1.50} & 1.27 \\ \cline{2-2}
 & V-04 & 15.70 & 172.81 & 29.41 & 46.65 & 2.07 & \textbf{1.33} & 1.45 & 29.91 & 26.18 & 5.96 & 330.20 & 74.53 & 0.03 & 71.71 & 51.40 & \textbf{61.43} & \textbf{1.47} & 1.45 \\ \cline{1-20}
\shortstack{Factory Floor (2024-04)} & 2025-01 & 76.31 & 0.54 & 0.63 & 78.62 & 0.53 & 0.65 & \textbf{93.61} & 0.49 & 0.72 & 88.75 & \textbf{0.13} & 0.38 & 6.09 & 18.35 & 52.65 & 92.92 & 0.17 & \textbf{0.30} \\ \cline{1-20}
\end{tabular}

}
\end{table*}
\subsection{Global Localization Performance}
To demonstrate the effectiveness of our method, we conducted experiments on real-world datasets.

\smallskip

\noindent\textbf{Experimental Setup.}
We evaluated our approach across diverse environments (see datasets Section~\ref{sec:datasets}). Unlike prior works that often use multiple sequences from the same environment as reference, we restrict each dataset to a single reference sequence, making global localization more challenging and representative of practical, low-overlap conditions. For MCD \cite{mcd_dataset}, we trained on ntu\_day\_01 and evaluated on ntu\_day\_02, ntu\_day\_10, ntu\_night\_04, ntu\_night\_08, and ntu\_night\_13 using Ouster OS1-128 data. For NCLT \cite{nclt_dataset}, we used 2012-01-15 as reference and evaluated on seven later sequences. 
For Wild-Places \cite{wild_places}, we used V-03 as reference and V-01, V-02, and V-04 for evaluation. For our Factory Floor dataset, we used 2024-04 as reference and 2025-01 for evaluation.
We benchmark against BEVPlace++~\cite{luo2024bevplaceplusplus}, a state-of-the-art retrieval-based method with local alignment; LightLoc~\cite{li2025lightloc}, a state-of-the-art scene-coordinate regression approach; KISS-Matcher~\cite{lim2024kiss}, a recent feature-based point cloud registration method; and PosePN++~\cite{YU2022108685}, an absolute pose-regression model. BEVPlace++ we tested once trained on KITTI \cite{Geiger2012CVPR}, denoted by (K), and once trained on the reference sequence of the respective dataset, denoted by (S).
Since we were unable to achieve reasonable results with the default parameter settings in LightLoc, we tried different parameter settings for each dataset and reported the best results. For the KISS-Matcher, we used a grid size of 0.3\,m for the indoor Factory Floor dataset, while we used a grid size of 1\,m for all other outdoor scenes. For all evaluations involving the KISS-Matcher, a scan-to-map alignment was performed. For the Wild Places and Factory Floor datasets, the map was constructed as a global point cloud from keyframes. In the case of the MCD dataset, we used the provided terrestrial laser scans as the map. Due to the high level of noise in the keyframe-based map for the NCLT dataset, we excluded it from the evaluation. PosePN++ was trained with the default settings.
While BEVPlace++ and our method only estimate 3 DoF poses, LightLoc, KISS-Matcher and PosePN++ provide 6 DoF poses. In the evaluation, we only considered the 3 DoF errors for all methods.

\smallskip

\noindent\textbf{Results.}
Table~\ref{tab:results} summarizes the quantitative comparison across the
datasets. Our method outperforms all competing approaches on 11 out of 16 sequences in terms of success rate (SR), demonstrating its superior robustness in localization. In contrast, the median translational (TE) and rotational (RE) errors of our method are comparable to those of other approaches, and in some cases slightly worse.
In Fig.~\ref{fig:day_10_hor_plot}, we visualize the inlier poses from sequence ntu\_day\_10 of the MCD dataset for the respective methods, displayed in the horizontal plane. Here it can be seen that our method has better inlier coverage, especially in areas far from the reference trajectory.
Figure~\ref{fig:sr_vs_distance} further analyzes SR as a function of the distance to the closest reference pose. The competing methods degrade sharply with increasing distance, while our method maintains high SR even under low-overlap conditions. This indicates that our approach provides a clear advantage in challenging deployment scenarios where training and test data are less aligned.
\section{Conclusion}
We propose a LiDAR-based global localization method that outperforms prior methods in success rate on most evaluated sequences. A novel consistency loss enables self-supervised landmark learning, automatically discovering salient structures without explicit definitions. Future work will explore extending this approach to camera images and assessing its potential for scene-agnostic keypoint detection.
\clearpage
\appendix
\section*{Appendix}
In this supplementary material, we provide additional details on the proposed method. We describe how the landmark density is determined, outline the training procedure, present the learned landmarks across different datasets, and discuss failure cases and practical considerations for deployment.
\section{Landmark Density}
\begin{figure}[h]
  \centering
  \includesvg[width=0.49\textwidth]{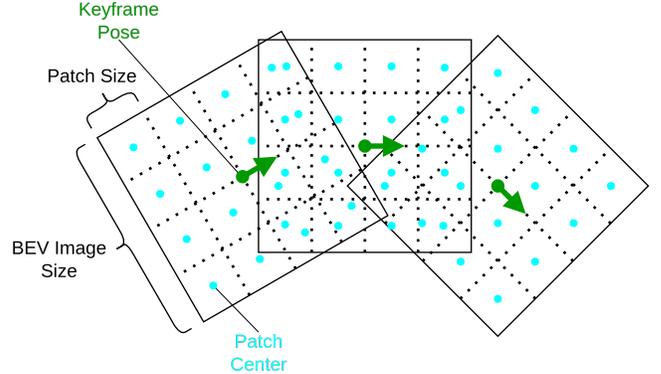} 
  \caption{Landmark initialization visualized for three exemplary keyframe poses.}
  \label{fig:lm_init}
\end{figure}

As stated in the main paper, the number of landmarks remains constant during training. Thus, the landmark density is determined when the initial landmarks are created. Landmark initialization consists of two steps. First, we iterate over all keyframe poses of the reference sequence. For each pose, we divide the area covered by the corresponding BEV image into $d_P \times d_P$ patches, matching the number of patches used in the landmark learning loss function, and place landmark candidates at the patch centers. This process is illustrated in Fig.~\ref{fig:lm_init}. Since the BEV images of neighboring keyframes overlap substantially, the accumulated landmark candidates have a much higher density than desired. Therefore, we apply a grid averaging filter with grid size $s_{grid}$ to reduce the number of landmarks. The final number of landmarks is thus controlled by the grid size $s_{grid}$, with at most one landmark retained per grid cell of area $a_{grid}=s^2_{grid}$. Instead of using the grid size $s_{grid}$ directly as a hyperparameter, we introduce the landmark density $\rho_{lm}$, defined as the ratio between the patch area $a_{patch}=l^2_{patch}$ and the grid area $a_{grid}$, where $l_{patch}$ denotes the patch edge length. This allows the grid size to adapt automatically to other hyperparameters, such as the BEV image resolution or patch size, while preserving a defined landmark density.

\begin{equation}
\rho_{lm}=\frac{a_{patch}}{a_{grid}}=\frac{l^2_{patch}}{s^2_{grid}}
\end{equation}

To be able to calculate the grid size as a function of the landmark density $\rho_{lm}$, we reformulate the formula:

\begin{equation}
s_{grid} =\frac{l_{patch}}{\sqrt{\rho_{lm}}}
\end{equation}

We can now use $\rho_{lm}$ as a control parameter and adaptively compute the grid size used to downsample the initial landmark candidates. For example, when $\rho_{lm}=1.0$, the grid size matches the patch size. Empirically, however, we found that a lower landmark density leads to better results, and therefore we use $\rho_{lm}=0.2$ in all experiments. This means that the grid size $s_{grid}$ is approximately twice the patch edge length $l_{patch}$. Although this introduces a relative sparsity of landmarks during landmark position learning, it ultimately improves localization performance.

\section{Training Setup}
We implemented the proposed method in PyTorch \cite{paszke2019pytorch}. We set the hyperparameter $d_P$ to 16, which yields $16 \times 16$ patches in the loss function (see main paper Fig.\,2). The loss weights are chosen as $\alpha = 1.0$, $\beta = 30.0$ and $\gamma = 3.0$. Optimization was performed using stochastic gradient descent (SGD) with a momentum of 0.9, combined with a StepLR scheduler. The learning rate was initialized at $4\times10^{-4}$ and decayed to $4\times10^{-5}$. For all experiments, 3\,\% of the reference sequence frames were used for validation.

\section{Landmark Analysis on Datasets}
\begin{figure*}[b]
\centering
\includesvg[width=0.99\textwidth, height=0.6\textheight]{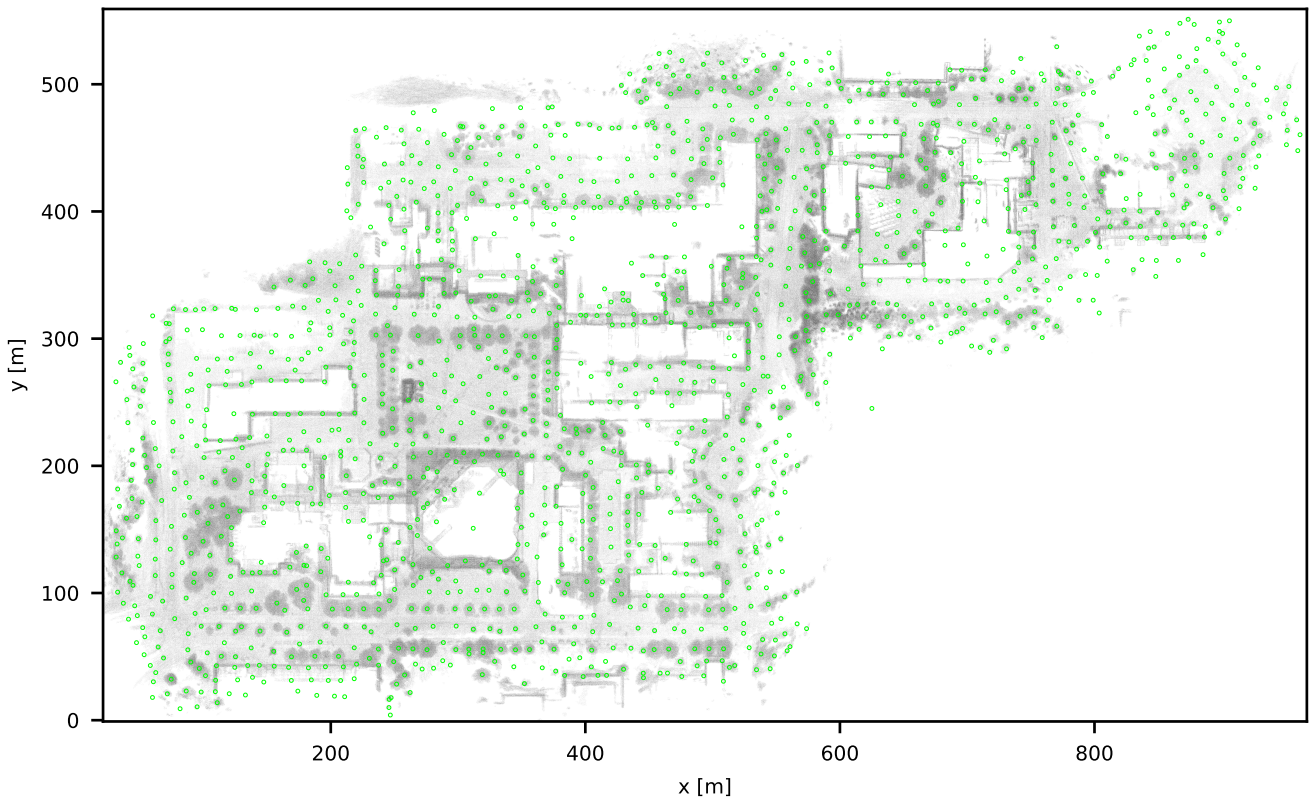}
\caption{Landmarks in \textbf{green} after self-supervised joint landmark detection and positioning on sequence 2012-01-15 of the NCLT dataset \cite{nclt_dataset} in a campus environment. In total, 1592 landmarks are utilized in this scene. A point cloud of the entire scene in the background is used for illustration purposes.}
\label{fig:lms_nclt}
\end{figure*}

\begin{figure*}[t]
\centering
\includesvg[width=0.85\textwidth, height=0.6\textheight]{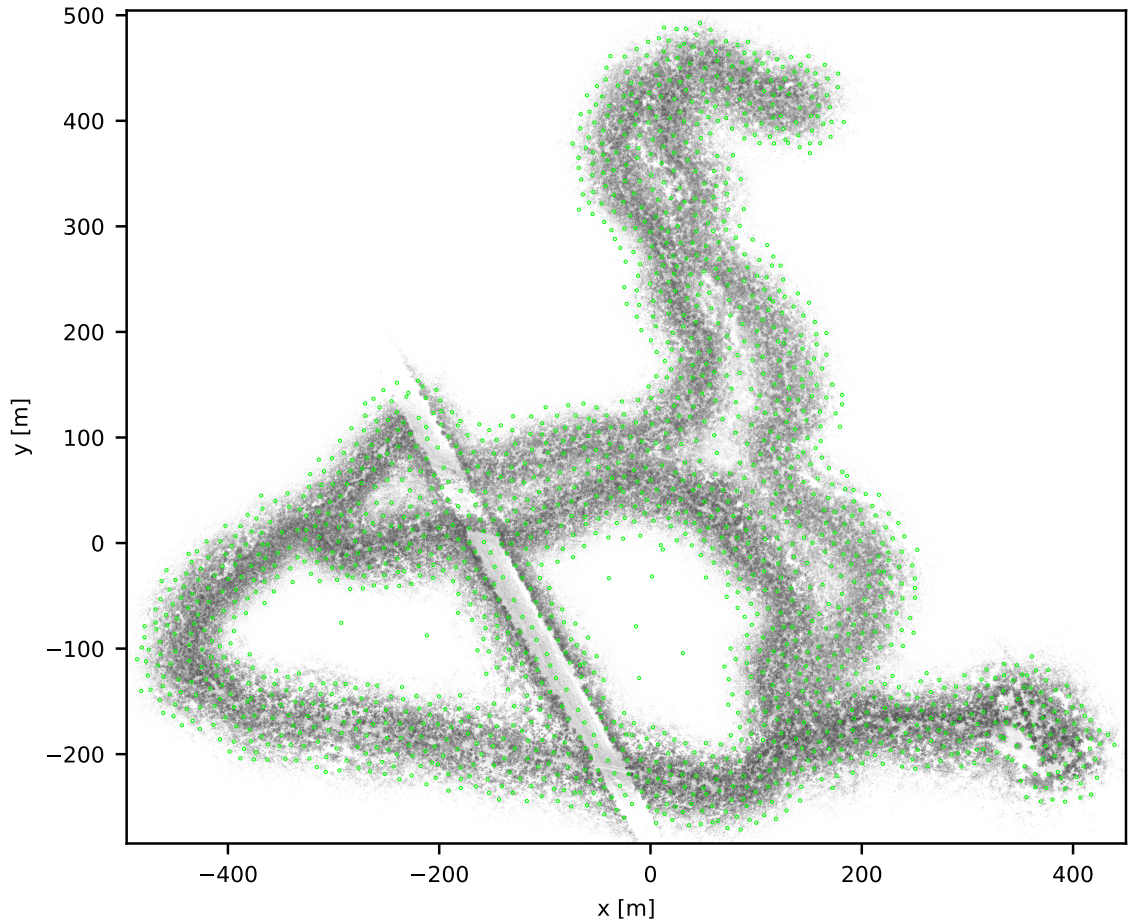}
\caption{Landmarks in \textbf{green} after self-supervised joint landmark detection and positioning on sequence V-03 of the Wild-Places dataset \cite{wild_places} in a forest environment. In total, 1757 landmarks are utilized in this scene. A point cloud of the entire scene in the background is used for illustration purposes.}
\label{fig:lms_wp}
\end{figure*}

In Figs.~\ref{fig:teaser}, \ref{fig:lms_nclt}, and \ref{fig:lms_wp}, we illustrate the learned landmarks for each dataset used for localization. For the MCD dataset \cite{mcd_dataset} and the NCLT dataset \cite{nclt_dataset}, the method selects tree trunks and building corners among other things as landmarks. In contrast, for the Wild-Places dataset \cite{wild_places}, it is challenging even for a human observer to provide an intuitive explanation for the landmark locations.

\section{Failure Cases}
In our experiments, suboptimal parameter settings degrade performance rather than cause complete failure. A low landmark density, $\rho_{lm} < 0.1$, may lead to landmark-free regions, whereas a high landmark density, $\rho_{lm} > 0.5$, may cause landmark collapse (co-located landmarks). While the latter does not affect the heatmaps, it can destabilize the correspondence loss. Performance degradation can also occur when training is stopped too early. In addition, the proposed method may fail in structureless or ambiguous scenes.
\section{Practical Considerations for Deployment}
In this section, we discuss practical considerations for deploying the proposed method in real-world robotic applications.

\subsection{Gravity Alignment}
LiDAR point clouds may require gravity alignment, particularly in non-planar environments. This can be achieved using IMU-based attitude estimation, which also enables 5 DoF pose estimation when combined with our method, which estimates 3 DoF.

\subsection{Motion Distortion Correction}
LiDAR point clouds recorded on dynamic platforms may capture the environment with motion-induced distortions. These distortions should be corrected, for example, by combining IMU angular velocity measurements with wheel-odometry-based velocity estimates.
\subsection{Continuous Localization and Sensor Fusion}
The proposed method focuses on LiDAR-based global localization. In our experiments, the global pose is estimated in a one-shot manner without prior information. For real-world deployment, however, sequential global localization estimates can be fused with IMU data and wheel odometry to enable continuous localization with improved robustness. In this setting, an additional inlier threshold may help reject global pose estimates supported by too few landmark inliers.

\subsection{Incremental Map Updates}
The current framework does not directly support incremental map updates. However, a simple extension would be to initialize the network and the landmark list with additional placeholder landmarks. These placeholder landmarks could be masked in the loss and reinitialized when the map expands, followed by a fine-tuning stage. Since the network uses LeakyReLU activations, initially inactive correspondence outputs remain reactivatable. Overall, this extension could be realized with modest effort.
\clearpage

{
    \small
    \bibliographystyle{ieeenat_fullname}
    \bibliography{main}
}


\end{document}

%% file: main.bib
@String(CVPR= {IEEE Conf. Comput. Vis. Pattern Recog.})

@String(CVPR  = {CVPR})

@inproceedings{do2022learning,
  title={Learning to detect scene landmarks for camera localization},
  author={Do, Tien and Miksik, Ondrej and DeGol, Joseph and Park, Hyun Soo and Sinha, Sudipta N},
  booktitle={Proceedings of the IEEE/CVF Conference on Computer Vision and Pattern Recognition},
  pages={11132--11142},
  year={2022}
}

@inproceedings{do2024improved,
  title={Improved scene landmark detection for camera localization},
  author={Do, Tien and Sinha, Sudipta N},
  booktitle={2024 International Conference on 3D Vision (3DV)},
  pages={975--984},
  year={2024},
  organization={IEEE}
}

@InProceedings{mcd_dataset,
    author    = {Nguyen, Thien-Minh and Yuan, Shenghai and Nguyen, Thien Hoang and Yin, Pengyu and Cao, Haozhi and Xie, Lihua and Wozniak, Maciej and Jensfelt, Patric and Thiel, Marko and Ziegenbein, Justin and Blunder, Noel},
    title     = {MCD: Diverse Large-Scale Multi-Campus Dataset for Robot Perception},
    booktitle = {Proceedings of the IEEE/CVF Conference on Computer Vision and Pattern Recognition (CVPR)},
    month     = {June},
    year      = {2024},
    pages     = {22304-22313}
}

@ARTICLE { nclt_dataset,
    AUTHOR = { Nicholas Carlevaris-Bianco and Arash K. Ushani and Ryan M. Eustice },
    TITLE = { University of {Michigan} {North} {Campus} long-term vision and lidar dataset },
    JOURNAL = { International Journal of Robotics Research },
    YEAR = { 2015 },
    VOLUME = { 35 },
    NUMBER = { 9 },
    PAGES = { 1023--1035 },
}

@INPROCEEDINGS{wild_places,

  author={Knights, Joshua and Vidanapathirana, Kavisha and Ramezani, Milad and Sridharan, Sridha and Fookes, Clinton and Moghadam, Peyman},

  booktitle={2023 IEEE International Conference on Robotics and Automation (ICRA)}, 

  title={Wild-Places: A Large-Scale Dataset for Lidar Place Recognition in Unstructured Natural Environments}, 

  year={2023},

  volume={},

  number={},

  pages={11322-11328},

  doi={10.1109/ICRA48891.2023.10160432}}

@ARTICLE{10494918,

  author={Shi, Chenghao and Chen, Xieyuanli and Xiao, Junhao and Dai, Bin and Lu, Huimin},

  journal={IEEE Transactions on Robotics}, 

  title={Fast and Accurate Deep Loop Closing and Relocalization for Reliable LiDAR SLAM}, 

  year={2024},

  volume={40},

  number={},

  pages={2620-2640},

  doi={10.1109/TRO.2024.3386363}}

@ARTICLE{luo2024bevplaceplusplus,
  author={Luo, Lun and Cao, Si-Yuan and Li, Xiaorui and Xu, Jintao and Ai, Rui and Yu, Zhu and Chen, Xieyuanli},
  journal={IEEE Transactions on Robotics}, 
  title={BEVPlace++: Fast, Robust, and Lightweight LiDAR Global Localization for Autonomous Ground Vehicles}, 
  year={2025},
  volume={41},
  number={},
  pages={4479-4498},
  doi={10.1109/TRO.2025.3585385}}

@INPROCEEDINGS{bonnbev,

  author={Gupta, Saurabh and Guadagnino, Tiziano and Mersch, Benedikt and Vizzo, Ignacio and Stachniss, Cyrill},

  booktitle={2024 IEEE International Conference on Robotics and Automation (ICRA)}, 

  title={Effectively Detecting Loop Closures using Point Cloud Density Maps}, 

  year={2024},

  volume={},

  number={},

  pages={10260-10266},

  doi={10.1109/ICRA57147.2024.10610962}}

@InProceedings{sgloc,
    author    = {Li, Wen and Yu, Shangshu and Wang, Cheng and Hu, Guosheng and Shen, Siqi and Wen, Chenglu},
    title     = {SGLoc: Scene Geometry Encoding for Outdoor LiDAR Localization},
    booktitle = {Proceedings of the IEEE/CVF Conference on Computer Vision and Pattern Recognition (CVPR)},
    month     = {June},
    year      = {2023},
    pages     = {9286-9295}
}

@InProceedings{Yang_2024_CVPR,
    author    = {Yang, Bochun and Li, Zijun and Li, Wen and Cai, Zhipeng and Wen, Chenglu and Zang, Yu and Muller, Matthias and Wang, Cheng},
    title     = {LiSA: LiDAR Localization with Semantic Awareness},
    booktitle = {Proceedings of the IEEE/CVF Conference on Computer Vision and Pattern Recognition (CVPR)},
    month     = {June},
    year      = {2024},
    pages     = {15271-15280}
}

@article{van2014scikit,
  title={scikit-image: image processing in Python},
  author={Van der Walt, Stefan and Sch{\"o}nberger, Johannes L and Nunez-Iglesias, Juan and Boulogne, Fran{\c{c}}ois and Warner, Joshua D and Yager, Neil and Gouillart, Emmanuelle and Yu, Tony},
  journal={PeerJ},
  volume={2},
  pages={e453},
  year={2014},
  publisher={PeerJ Inc.}
}

@article{fischler1981random,
  title={Random sample consensus: a paradigm for model fitting with applications to image analysis and automated cartography},
  author={Fischler, Martin A and Bolles, Robert C},
  journal={Communications of the ACM},
  volume={24},
  number={6},
  pages={381--395},
  year={1981},
  publisher={ACM New York, NY, USA}
}

@INPROCEEDINGS{lim2024kiss,

  author={Lim, Hyungtae and Kim, Daebeom and Shin, Gunhee and Shi, Jingnan and Vizzo, Ignacio and Myung, Hyun and Park, Jaesik and Carlone, Luca},

  booktitle={2025 IEEE International Conference on Robotics and Automation (ICRA)}, 

  title={KISS-Matcher: Fast and Robust Point Cloud Registration Revisited}, 

  year={2025},

  volume={},

  number={},

  pages={11104-11111},

  doi={10.1109/ICRA55743.2025.11127458}}

@inproceedings{seferbekov2018feature,
  title={Feature pyramid network for multi-class land segmentation},
  author={Seferbekov, Selim and Iglovikov, Vladimir and Buslaev, Alexander and Shvets, Alexey},
  booktitle={Proceedings of the IEEE conference on computer vision and pattern recognition workshops},
  pages={272--275},
  year={2018}
}

@INPROCEEDINGS{10610818,
author={Skuddis, David and Haala, Norbert},
booktitle={2024 IEEE International Conference on Robotics and Automation (ICRA)}, 
title={DMSA - Dense Multi Scan Adjustment for LiDAR Inertial Odometry and Global Optimization}, 
year={2024},
volume={},
number={},
pages={12027-12033},
doi={10.1109/ICRA57147.2024.10610818}}

@INPROCEEDINGS{6751291,

  author={Yang, Jiaolong and Li, Hongdong and Jia, Yunde},

  booktitle={2013 IEEE International Conference on Computer Vision}, 

  title={Go-ICP: Solving 3D Registration Efficiently and Globally Optimally}, 

  year={2013},

  volume={},

  number={},

  pages={1457-1464},

  doi={10.1109/ICCV.2013.184}}

@INPROCEEDINGS{5152473,

  author={Rusu, Radu Bogdan and Blodow, Nico and Beetz, Michael},

  booktitle={2009 IEEE International Conference on Robotics and Automation}, 

  title={Fast Point Feature Histograms (FPFH) for 3D registration}, 

  year={2009},

  volume={},

  number={},

  pages={3212-3217},

  doi={10.1109/ROBOT.2009.5152473}}

@article{YU2022108685,
title = {LiDAR-based localization using universal encoding and memory-aware regression},
journal = {Pattern Recognition},
volume = {128},
pages = {108685},
year = {2022},
issn = {0031-3203},
doi = {https://doi.org/10.1016/j.patcog.2022.108685},
url = {https://www.sciencedirect.com/science/article/pii/S0031320322001662},
author = {Shangshu Yu and Cheng Wang and Chenglu Wen and Ming Cheng and Minghao Liu and Zhihong Zhang and Xin Li},
}

@INPROCEEDINGS{10657300,

  author={Li, Wen and Yang, Yuyang and Yu, Shangshu and Hu, Guosheng and Wen, Chenglu and Cheng, Ming and Wang, Cheng},

  booktitle={2024 IEEE/CVF Conference on Computer Vision and Pattern Recognition (CVPR)}, 

  title={DiffLoc: Diffusion Model for Outdoor LiDAR Localization}, 

  year={2024},

  volume={},

  number={},

  pages={15045-15054},

  doi={10.1109/CVPR52733.2024.01425}}

@INPROCEEDINGS{10203089,

  author={Wang, Sijie and Kang, Qiyu and She, Rui and Wang, Wei and Zhao, Kai and Song, Yang and Tay, Wee Peng},

  booktitle={2023 IEEE/CVF Conference on Computer Vision and Pattern Recognition (CVPR)}, 

  title={HypLiLoc: Towards Effective LiDAR Pose Regression with Hyperbolic Fusion}, 

  year={2023},

  volume={},

  number={},

  pages={5176-5185},

  doi={10.1109/CVPR52729.2023.00501}}

@INPROCEEDINGS{8593953,

  author={Kim, Giseop and Kim, Ayoung},

  booktitle={2018 IEEE/RSJ International Conference on Intelligent Robots and Systems (IROS)}, 

  title={Scan Context: Egocentric Spatial Descriptor for Place Recognition Within 3D Point Cloud Map}, 

  year={2018},

  volume={},

  number={},

  pages={4802-4809},

  doi={10.1109/IROS.2018.8593953}}

@ARTICLE{9610172,

  author={Kim, Giseop and Choi, Sunwook and Kim, Ayoung},

  journal={IEEE Transactions on Robotics}, 

  title={Scan Context++: Structural Place Recognition Robust to Rotation and Lateral Variations in Urban Environments}, 

  year={2022},

  volume={38},

  number={3},

  pages={1856-1874},

  doi={10.1109/TRO.2021.3116424}}

@inproceedings{chen2020rss, 
		author = {X. Chen and T. L\"abe and A. Milioto and T. R\"ohling and O. Vysotska and A. Haag and J. Behley and C. Stachniss},
		title  = {{OverlapNet: Loop Closing for LiDAR-based SLAM}},
		booktitle = {Proceedings of Robotics: Science and Systems (RSS)},
		year = {2020}
}

@INPROCEEDINGS{8578568,

  author={Uy, Mikaela Angelina and Lee, Gim Hee},

  booktitle={2018 IEEE/CVF Conference on Computer Vision and Pattern Recognition}, 

  title={PointNetVLAD: Deep Point Cloud Based Retrieval for Large-Scale Place Recognition}, 

  year={2018},

  volume={},

  number={},

  pages={4470-4479},

  doi={10.1109/CVPR.2018.00470}}

@ARTICLE{ma2022ral,
  author={Ma, Junyi and Zhang, Jun and Xu, Jintao and Ai, Rui and Gu, Weihao and Chen, Xieyuanli},
  journal={IEEE Robotics and Automation Letters}, 
  title={OverlapTransformer: An Efficient and Yaw-Angle-Invariant Transformer Network for LiDAR-Based Place Recognition}, 
  year={2022},
  volume={7},
  number={3},
  pages={6958-6965},
  doi={10.1109/LRA.2022.3178797}}

@ARTICLE{cattaneo2022tro,
  author={Cattaneo, Daniele and Vaghi, Matteo and Valada, Abhinav},
  journal={IEEE Transactions on Robotics}, 
  title={LCDNet: Deep Loop Closure Detection and Point Cloud Registration for LiDAR SLAM}, 
  year={2022},
  volume={38},
  number={4},
  pages={2074-2093},
  doi={10.1109/TRO.2022.3150683}
 }

@ARTICLE{9645340,
author={Komorowski, Jacek and Wysoczanska, Monika and Trzcinski, Tomasz},
journal={IEEE Robotics and Automation Letters}, 
title={EgoNN: Egocentric Neural Network for Point Cloud Based 6DoF Relocalization at the City Scale}, 
year={2022},
volume={7},
number={2},
pages={722-729},
doi={10.1109/LRA.2021.3133593}}

@INPROCEEDINGS{10610810,

  author={Aoki, Koki and Koide, Kenji and Oishi, Shuji and Yokozuka, Masashi and Banno, Atsuhiko and Meguro, Junichi},

  booktitle={2024 IEEE International Conference on Robotics and Automation (ICRA)}, 

  title={3D-BBS: Global Localization for 3D Point Cloud Scan Matching Using Branch-and-Bound Algorithm}, 

  year={2024},

  volume={},

  number={},

  pages={1796-1802},

  doi={10.1109/ICRA57147.2024.10610810}}

@INPROCEEDINGS{dellenbach2021cticp,

  author={Dellenbach, Pierre and Deschaud, Jean-Emmanuel and Jacquet, Bastien and Goulette, François},

  booktitle={2022 International Conference on Robotics and Automation (ICRA)}, 

  title={CT-ICP: Real-time Elastic LiDAR Odometry with Loop Closure}, 

  year={2022},

  volume={},

  number={},

  pages={5580-5586},

  doi={10.1109/ICRA46639.2022.9811849}}

@inproceedings{li2025lightloc,
  title={LightLoc: Learning Outdoor LiDAR Localization at Light Speed},
  author={Li, Wen and Liu, Chen and Yu, Shangshu and Liu, Dunqiang and Zhou, Yin and Shen, Siqi and Wen, Chenglu and Wang, Cheng},
  booktitle={Proceedings of the Computer Vision and Pattern Recognition Conference},
  pages={6680--6689},
  year={2025}
}

@inproceedings{Geiger2012CVPR,
  author = {Andreas Geiger and Philip Lenz and Raquel Urtasun},
  title = {Are we ready for Autonomous Driving? The KITTI Vision Benchmark Suite},
  booktitle = {Conference on Computer Vision and Pattern Recognition (CVPR)},
  year = {2012}
}

@INPROCEEDINGS{res_block,

  author={He, Kaiming and Zhang, Xiangyu and Ren, Shaoqing and Sun, Jian},

  booktitle={2016 IEEE Conference on Computer Vision and Pattern Recognition (CVPR)}, 

  title={Deep Residual Learning for Image Recognition}, 

  year={2016},

  volume={},

  number={},

  pages={770-778},

  doi={10.1109/CVPR.2016.90}}

@article{paszke2019pytorch,
  title={Pytorch: An imperative style, high-performance deep learning library},
  author={Paszke, Adam and Gross, Sam and Massa, Francisco and Lerer, Adam and Bradbury, James and Chanan, Gregory and Killeen, Trevor and Lin, Zeming and Gimelshein, Natalia and Antiga, Luca and others},
  journal={Advances in neural information processing systems},
  volume={32},
  year={2019}
}

@ARTICLE{bow3d,
  author={Cui, Yunge and Chen, Xieyuanli and Zhang, Yinlong and Dong, Jiahua and Wu, Qingxiao and Zhu, Feng},
  journal={IEEE Robotics and Automation Letters}, 
  title={BoW3D: Bag of Words for Real-Time Loop Closing in 3D LiDAR SLAM}, 
  year={2023},
  volume={8},
  number={5},
  pages={2828-2835},
  doi={10.1109/LRA.2022.3221336}}

@ARTICLE{btc,

  author={Yuan, Chongjian and Lin, Jiarong and Liu, Zheng and Wei, Hairuo and Hong, Xiaoping and Zhang, Fu},

  journal={IEEE Transactions on Robotics}, 

  title={BTC: A Binary and Triangle Combined Descriptor for 3-D Place Recognition}, 

  year={2024},

  volume={40},

  number={},

  pages={1580-1599},

  doi={10.1109/TRO.2024.3353076}}

@INPROCEEDINGS{ gskim-2020-mulran, 

    TITLE={MulRan: Multimodal Range Dataset for Urban Place Recognition}, 

    AUTHOR={Giseop Kim and Yeong Sang Park and Younghun Cho and Jinyong Jeong and Ayoung Kim}, 

    BOOKTITLE = { Proceedings of the IEEE International Conference on Robotics and Automation (ICRA) },

    YEAR = { 2020 },

    MONTH = { May },

    ADDRESS = { Paris }

}

@INPROCEEDINGS{minkloc_3d,
  author={Komorowski, Jacek},
  booktitle={2021 IEEE Winter Conference on Applications of Computer Vision (WACV)}, 
  title={MinkLoc3D: Point Cloud Based Large-Scale Place Recognition}, 
  year={2021},
  volume={},
  number={},
  pages={1789-1798},
  doi={10.1109/WACV48630.2021.00183}}

@INPROCEEDINGS{vid2021locus,

  author={Vidanapathirana, Kavisha and Moghadam, Peyman and Harwood, Ben and Zhao, Muming and Sridharan, Sridha and Fookes, Clinton},

  booktitle={2021 IEEE International Conference on Robotics and Automation (ICRA)}, 

  title={Locus: LiDAR-based Place Recognition using Spatiotemporal Higher-Order Pooling}, 

  year={2021},

  volume={},

  number={},

  pages={5075-5081},

  doi={10.1109/ICRA48506.2021.9560915}}

@inproceedings{vid2022logg3d,
  title={LoGG3D-Net: Locally Guided Global Descriptor Learning for 3D Place Recognition},
  author={Vidanapathirana, Kavisha and Ramezani, Milad and Moghadam, Peyman and Sridharan, Sridha and Fookes, Clinton},
  booktitle={2022 International Conference on Robotics and Automation (ICRA)},
  pages={2215--2221},
  year={2022}
}
